\documentclass{svproc}
\usepackage{url}

\usepackage{graphicx}
\usepackage{booktabs}
\usepackage[caption=false]{subfig}
\usepackage[export]{adjustbox}
\usepackage{amsmath, amssymb}
\usepackage{enumitem}

\newcommand\lidar{LiDAR}

\begin{document}
\mainmatter
\title{Trustworthy Automated Driving through Qualitative Scene Understanding and Explanations\thanks{This work is funded by the European Commission through the AI4CCAM project under grant agreement No 101076911.}}
\toctitle{Trustworthy Automated Driving through Qualitative Scene Understanding and Explanations}
\titlerunning{Trustworthy AD through Qualitative Scene Understanding and Explanations}  %
\author{Nassim Belmecheri\inst{1} \and Arnaud Gotlieb\inst{1} \and Nadjib Lazaar\inst{2} \and Helge Spieker\inst{1}}
\institute{Simula Research Laboratory, Oslo, Norway\\\email{\{nassim,arnaud,helge\}@simula.no}
\and
LIRMM, University of Montpellier, CNRS, Montpellier, France\\\email{nadjib.lazaar@lirmm.fr}}

\maketitle

\begin{abstract}
We present the Qualitative Explainable Graph (QXG): a unified symbolic and qualitative representation for scene understanding in urban mobility. QXG enables the interpretation of an automated vehicle's environment using sensor data and machine learning models. It leverages spatio-temporal graphs and qualitative constraints to extract scene semantics from raw sensor inputs, such as LiDAR and camera data, offering an intelligible scene model. Crucially, QXG can be incrementally constructed in real-time, making it a versatile tool for in-vehicle explanations and real-time decision-making across various sensor types. Our research showcases the transformative potential of QXG, particularly in the context of automated driving, where it elucidates decision rationales by linking the graph with vehicle actions. These explanations serve diverse purposes, from informing passengers and alerting vulnerable road users (VRUs) to enabling post-analysis of prior behaviours.

\end{abstract}

\section{Introduction}

Artificial intelligence (AI) methods are at the core of automated driving (AD) and connected mobility.
However, passing control to an AI-based system and trusting its decisions requires the ability to request explanations for these decisions \cite{omeiza_explanations_2022}.
In fact, societal acceptance of AD significantly depends on these AI models' trustworthiness, transparency and reliability \cite{nastjuk_what_2020}.
Still, this is an open challenge as many of the state-of-the-art deep learning (DL) models are opaque and not inherently explainable by themselves \cite{atakishiyev_explainable_2023}.

In recent years, a number of Explainable AI (XAI) methods with a focus on automated driving have been proposed.
Following \cite{atakishiyev_explainable_2023}, they fall into three main categories: 
a) \emph{Vision-based XAI} related to highlighting the area of an image that influences a perception model towards a certain output \cite{omeiza_explanations_2022}; 
b) \emph{Feature-based importance scores} quantify the influence of each individual input feature on the model output; and 
c) \emph{Textual-based XAI} that aims to formulate explanations as intelligible arguments using natural language processing \cite{kim_toward_2021}.
Unfortunately, automated support for multi-sensor and video-based scene explanation is still restricted to quantitative analysis, e.g., saliency heatmaps \cite{omeiza_explanations_2022}.

In this work, we approach qualitative methods for scene understanding by using \emph{Qualitative Explainable Graphs (QXG)} and, based on this representation, a novel method for action explanation.
A QXG captures the spatio-temporal dynamics of a scene via qualitative algebras, i.e., a description of the relative positions (e.g., pedestrian north of ego car), a qualitative distance (e.g., pedestrian far from ego car) and their direction towards each other (e.g., ego car approaching static pedestrian).
From these graphs, interpretable machine learning models are trained to provide justification for taken actions.
Our results on the real-world nuScenes dataset \cite{caesar_nuscenes_2020} show that the QXG can be efficiently constructed incrementally in real-time and that it serves to correctly explain actions.

\section{Background \& Related Work}

\paragraph{Qualitative Calculi}

A \emph{qualitative calculus (QC)} is a computational method for analyzing qualitative connections among physical attributes, such as position, velocity, and acceleration, independently of precise quantitative data \cite{dylla_survey_2015}. QC can be parameterized by a qualitative algebra tailored to temporal dynamics, spatial relationships, or a combination of both \cite{allen_maintaining_1983,renz_qualitative_2007}.
For automated driving, qualitative reasoning is utilized through ontologies \cite{westhofen_using_2022} and neurosymbolic online abduction \cite{suchan_commonsense_2021}. 
This enables encoding complex driving scenarios and traffic dynamics, especially when obtaining precise measurements is challenging or unfeasible.
QC are commonly used in spatio-temporal reasoning to describe the relations between sets of objects in a space or over time, e.g. the positioning, distance, or orientation of vulnerable road users (VRUs) in relation to a vehicle.

In this work, we rely on four qualitative calculi \cite{dylla_survey_2015} for all spatial aspects:
\begin{enumerate}[leftmargin=*]
\item \textbf{Qualitative Distance Calculus ($QDC$)} \cite{renz_qualitative_2007} focuses on representing and reasoning about distances between objects in a qualitative manner, without relying on precise metric measurements. 

\item \textbf{Rectangle Algebra ($RA$)} \cite{renz_qualitative_2007} provides a qualitative relative positioning of objects represented as rectangle rather than as points, i.e., involving spatial dimensions. It is a two-dimensional extension of the Allen's interval algebra \cite{allen_maintaining_1983} and valuable for describing object orientations and spatial relationships.

\item \textbf{Basic Qualitative Trajectory Calculus ($QTC_b$)} \cite{dylla_survey_2015} deals with qualitative representations of object trajectories and their interactions. It enables reasoning about the motion and paths of objects without the need for detailed numerical data. 
It shall be noted that the heading needs to be inferred temporally and is, unlike the other calculi, not a pure spatial relationship.

\item \textbf{Star Calculus} \cite{renz_qualitative_2007} is a qualitative calculus designed to represent and reason about spatial regions and their relationships. It is useful for describing regions of influence, zones, and coverage areas in automated driving scenarios. We apply $STAR_4$ which divides the surrounding into 4 quarters.
\end{enumerate}

\noindent\textit{Qualitative Scene Understanding}
Scene understanding involves gathering and organising spatial and temporal information regarding objects, including vehicles, VRUs, and static elements, across a sequence of frames~\cite{xue_survey_2018}. 
At its core, scene understanding encompasses perception tasks like object detection and image segmentation~\cite{muhammad_vision-based_2022}. 
In qualitative scene understanding, we operate at a higher level, focusing on the qualitative depiction of a scene, emphasising objects and their temporal and spatial relations.

In the context of automated driving, a scene is formally represented as a sequence of $n$ frames, depicted as $\mathcal{S}=\langle f_1,\ldots, f_n \rangle$. Object detection and tracking, as part of this process, involves detecting objects in a given frame $f_k$ within $\mathcal{S}$, determining their bounding boxes, and tracking their movement relative to previously detected objects in preceding frames.
We assume a set of $m$ detected objects, denoted as $\mathcal{O}=\{ o_1,\ldots o_m\}$, to be present in $\mathcal{S}$, where each object $o_i$ appears in at least one frame $f_j\in \mathcal{S}$.

Scene understanding primarily focuses on assessing the situational context. 
Its outcome can be used for decision-making, trajectory prediction, or providing explanations and analyses of various aspects within the scene.

\section{Qualitative Explainable Graph}

The Qualitative Explainable Graph (QXG) is a scene representation format describing qualitative spatial-temporal relations among objects in a scene\footnote{The QXG was first described in \cite{belmecheri_acquiring_2023}; in this work we extend it to a more complete representation with multiple qualitative calculi.}.
The graph representing a scene $S$ is composed of one node per object in $\mathcal{O}$ with edges $\mathcal{V}$ between objects that appear jointly in at least one frame $f$.

\begin{figure}[t]
    \centering
    \includegraphics[scale=0.3,width=\textwidth]{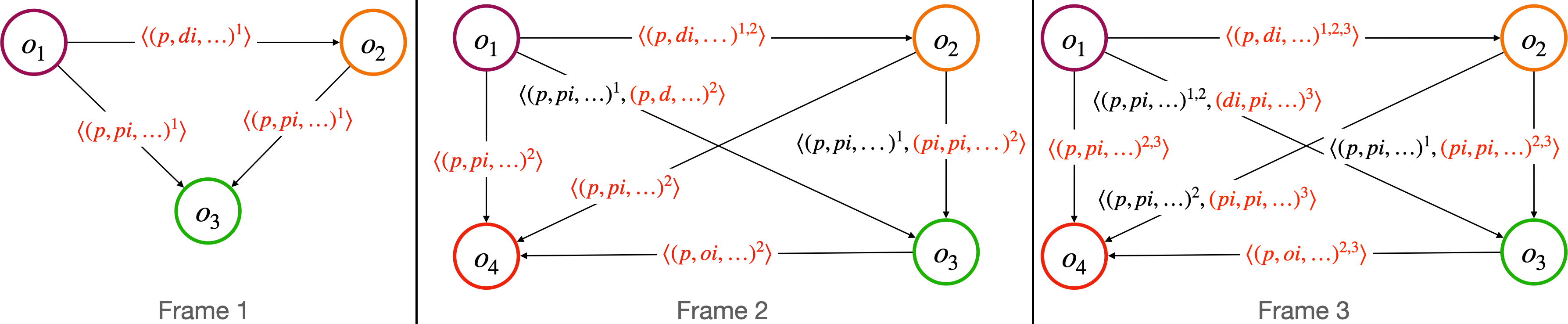}
    \caption{Illustration of the successive construction process of the QXG over multiple frames. For simplification, only the rectangle algebra relation is depicted.}
    \label{fig:qxgbuilder}
\end{figure}

The QXG captures the relation between two objects over the temporal course of a scene through spatial relations and their changes.
The selection of the spatial calculi is a parameter of the QXG and, depending on the needed granularity and or use case requirements, alternative calculi may be considered.
In the context of this work, we describe the relation between objects by a mixture of the four calculi mentioned above to capture the necessary spatial information.
Through the combination of these calculi, we cover the relevant aspects to understand and explain scenes through qualitative graphs.
Nevertheless, the formulation of the QXG and its usage is generally independent of the specific calculi chosen as long as they are expressive enough to cover at least the relative positioning and distance of objects, although extra calculi \cite{dylla_survey_2015} might be desirable, depending on the use case, to enrich the representation. 

\section{Action Explanation}

The QXG offers a unified qualitative representation of scenes and their object dynamics, enabling post-hoc explanations of actions taken by individual actors without recreation of the graph or retraining of the method. 
These explanations consist of object-pair relation chains that justify why an action was taken, i.e. a rationalization, considering an external perspective.

To approach action explanation, we frame the task as a one-against-all classification problem.
Our method is trained on a labelled dataset comprising QXGs and annotated actions, using the real-world nuScenes dataset \cite{caesar_nuscenes_2020} for QXG generation from \lidar{} data.
During training, annotated QXGs form a training dataset. 
For each action in a scene, we extract the $t$ most recent object-relation chains, creating joint feature vectors that describe the explanation context. 
These feature vectors are generated for the acting object and all other objects that appeared in the same frame in the last $t$ frames.
We train one-class classifiers for each action in the dataset. 
These classifiers assess the likelihood of a given object-pair relation chain causing a specific action against all other actions. 

During the explanation stage, we require a QXG and an action to be explained. 
Object-pairs involving the acting object are scored by the action's classifier. 
The pairs with the highest scores represent the most plausible explanation for the acting object's behaviour.
This approach allows us to flexibly adjust the explanation scope by altering the classifier score threshold. Additionally, by employing interpretable classifiers like tree-based models, we can provide decision paths as supplementary context for the highest classification scores.

\begin{figure}[t]
    \centering
    \subfloat[QXG action explanation process: From the action-annotated QXG the previous relations are extracted and classified according to the target action.\label{fig:actionexplanation}]{
    \includegraphics[width=0.65\textwidth,valign=b]{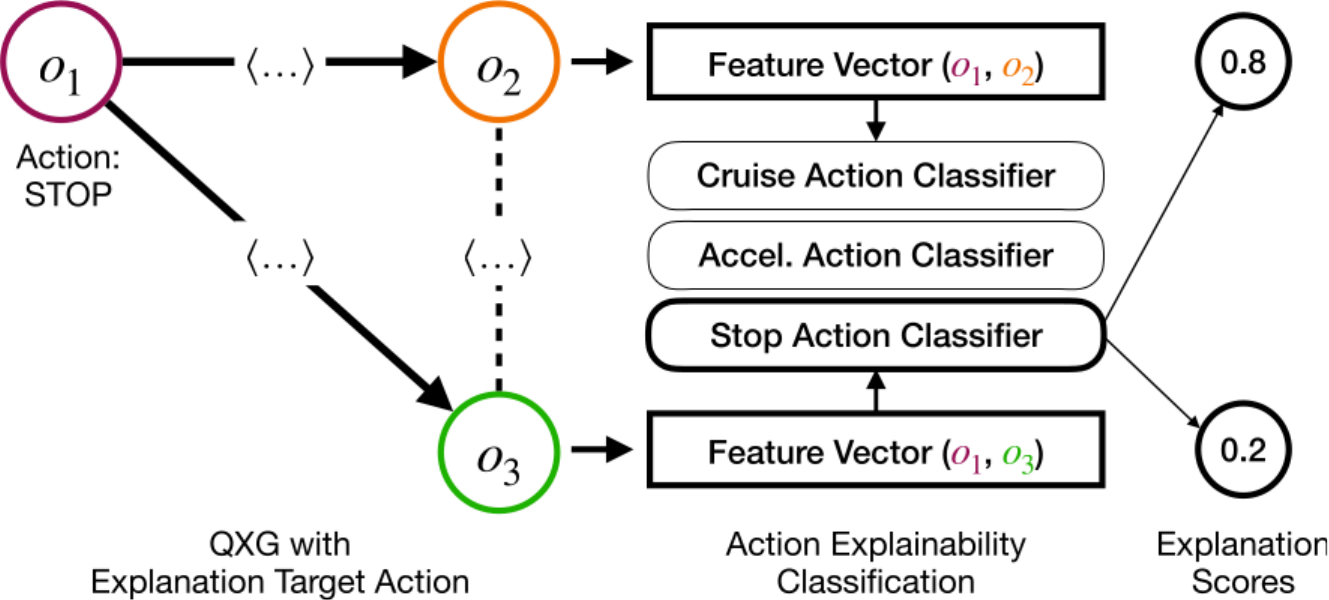}}\hfill%
    \subfloat[Classifier Results\label{tab:aeresults}]{
    \adjustbox{valign=b}{\begin{tabular}{lrr}
        \toprule
        Action      & Precision & Recall \\
        \midrule
        Cruising    & 89.7\,\% & 89.7\,\%\\
        Accel.      & 89.2\,\% & 89.2\,\%\\
        Stopping    & 90.6\,\% & 90.6\,\%\\
        \midrule
        Average     & 89.8\,\% & 89.8\,\%\\
        \bottomrule
    \end{tabular}}}
    \caption{An overview of the explanation process and results for the trained action explanation classifiers.}
\end{figure}

\section{Experiments}
We assess the action explanations using 850 QXG-represented scenes from the nuScenes dataset~\cite{caesar_nuscenes_2020}. The QXG is built incrementally, frame-by-frame from the top \lidar{} view, which is computationally efficient. Remarkably, even for frames with the maximum number of 160 objects, construction remains efficient, taking less than 50 milliseconds, thus achieving real-time QXG generation.

Random forests are trained as the interpretable action explainability classifiers on 595 scenes.
Our evaluation results on 255 held-out scenes, summarised in Table~\ref{tab:aeresults}, employ Precision and Recall as key metrics, gauging prediction correctness and sensitivity, respectively. 
Notably, Precision and Recall exhibit identical values, as we are dealing with a one-class classification scenario, and our test cases are abundant.

\begin{figure}[t]
    \centering
    \includegraphics[width=\textwidth]{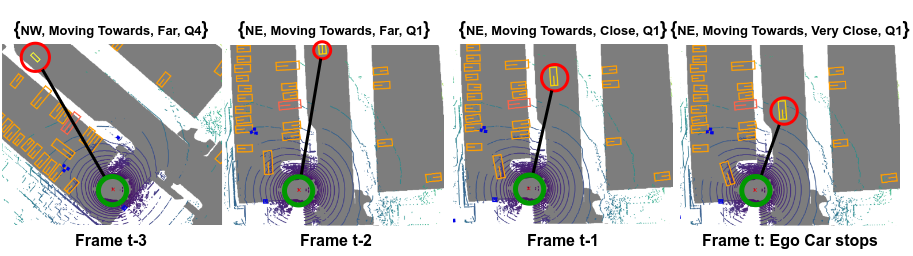}
    \caption{Example action explanation overlaid on the \lidar{} view: The car circled in red approaches the ego car, as captured by edge relations between these two objects above the images. Calculated from the specified calculi, the relations rationalise the stopping. NW: North west, NE: North east; Order of relations: $RA$, $QTC_b$, $QDC$, $STAR_4$.}
    \label{fig:exp_example}
\end{figure}

To provide an illustrative example, Figure~\ref{fig:exp_example} showcases an explanation for an ego car's decision to halt at a parking lot intersection, prompted by the approach of a yellow car.
While there are many objects (depicted in yellow), the action explanation correctly highlights the approaching car as the main incentive for the stopping maneuver.

\section{Conclusion}

Establishing a symbolic and qualitative comprehension of the vehicle's surroundings enhances communication not only with internal decision-making AI but also with other vehicles, VRUs, and external auditors, thereby bolstering system safety and reliability~\cite{llorca2021trustworthy}. In this paper, we have introduced the Qualitative Explainable Graph (QXG), which is a spatio-temporal representation of automated driving scenarios that can be constructed incrementally in real-time.
The key advantage of employing a qualitative scene representation lies in its capacity for introspection and in-depth analysis. 
Action explanations can be performed by training interpretable tree-based classifiers from QXGs and we showed that this can be efficiently performed on the real-world nuScenes automated driving dataset.
In future work, we will deepen the use of QXGs for AI in CCAM~\cite{llorca2021trustworthy}, such as vehicle-to-vehicle scene understanding, intelligible explanations for VRUs, and advanced message-passing techniques to enhance the action explanation process.
\bibliographystyle{splncs03}
\bibliography{refs}   %

\end{document}